\documentclass[11pt,a4paper]{article}
\usepackage[hyperref]{acl2018}
\usepackage{CJK}
\usepackage{times}
\usepackage{graphicx}
\usepackage{url}
\usepackage{textcomp}

\newcommand{\tabincell}[2]{\begin{tabular}{@{}#1@{}}#2\end{tabular}}  

\title{A Gap-Based Framework for Chinese Word Segmentation via\\ Very Deep Convolutional Networks}

\author{
  Zhiqing Sun \\
  Peking University \\\And
  Gehui Shen \\
  Peking University \\
  {\tt \{1500012783, jueliangguke, zhdeng\}@pku.edu.cn} \\\And
  Zhihong Deng \\
  Peking University \\}

\date{}

\begin{document}
\maketitle
\global\csname @topnum\endcsname 0
\global\csname @botnum\endcsname 0
\begin{abstract}
	Most previous approaches to Chinese word segmentation can be roughly classified into character-based and word-based methods. The former regards this task as a sequence-labeling problem, while the latter directly segments character sequence into words. However, if we consider segmenting a given sentence, the most intuitive idea is to predict whether to segment for each gap between two consecutive characters, which in comparison makes previous approaches seem too complex. Therefore, in this paper, we propose a gap-based framework to implement this intuitive idea. Moreover, very deep convolutional neural networks, namely, ResNets and DenseNets, are exploited in our experiments. Results show that our approach outperforms the best character-based and word-based methods on 5 benchmarks, without any further post-processing module (e.g. Conditional Random Fields) nor beam search.
\end{abstract}

\begin{figure}[t!]
\includegraphics[width=7.7 cm]{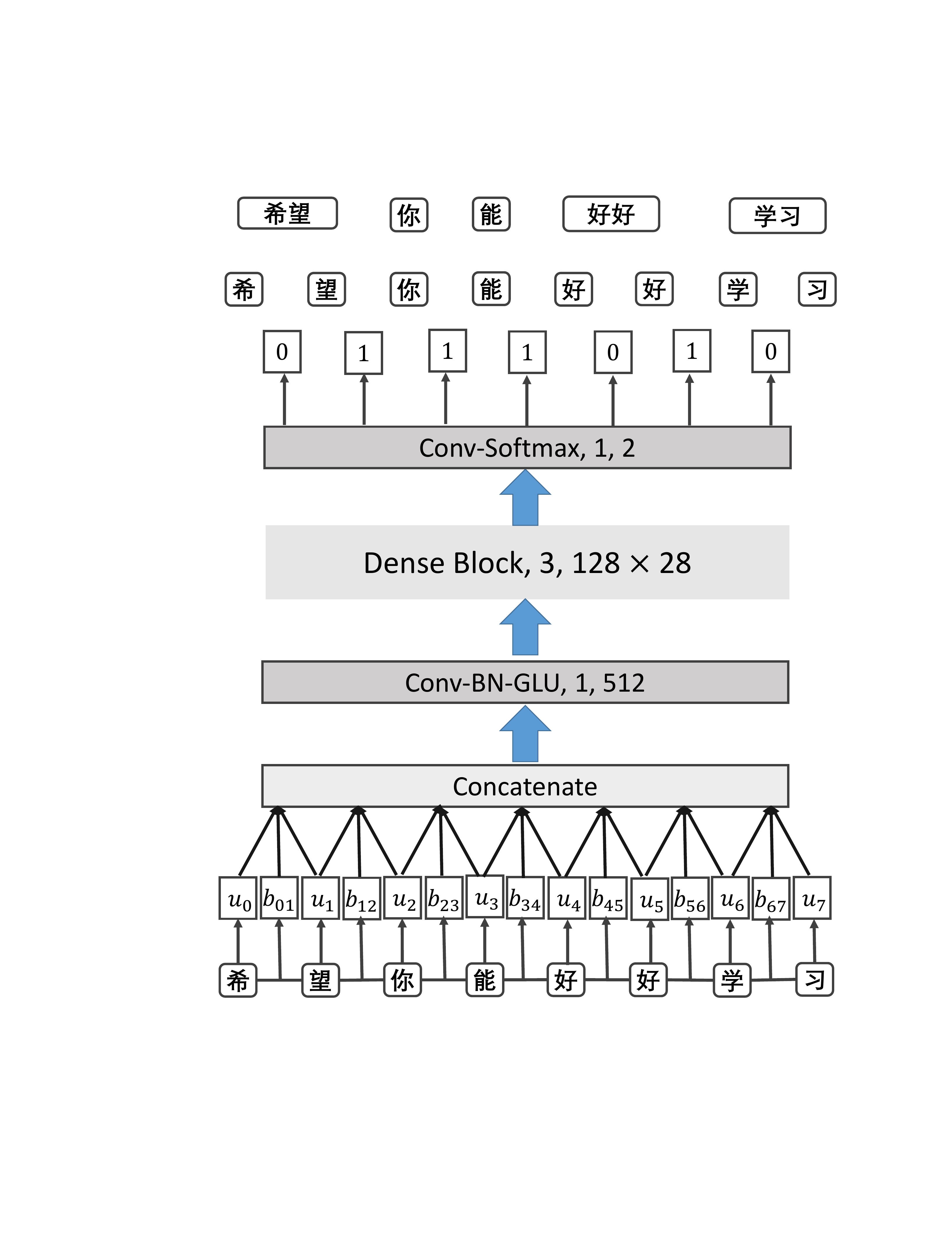}
 \begin{CJK}{UTF8}{gbsn}
\caption{A Gap-Based Convolutional Network with a dense block that directly segment ``希望你能好好学习'' into ``希望 (hope) \quad 你 (you) \quad 能 (can) \quad 好好 (happily) \quad 学习 (study) ''.}
\label{fig:framework}
\end{CJK}
\end{figure}

\section{Introduction}

	Unlike English, Chinese are written without explicit word delimiters, which makes word segmentation a fundamental and preliminary task in Chinese natural language processing. Recently, neural approaches for Chinese Word Segmentation (CWS) are attracting huge interest and a great deal of neural models have given competitive results to the best statistical models.

	Previous neural approaches to CWS can be roughly classified into character-based and word-based. The former regards this task as a sequence-labeling problem, while the latter directly segments character sequence into words.

	Since \citet{O03-4002}, most character-based methods use \{B, I, E, S\} labels to denote the Beginning, Internal, End of a word and a Single-character word, respectively. To get the label scores for each character, tensor neural network \citep{pei-ge-chang:2014:P14-1},  recursive neural network \citep{chen-EtAl:2015:ACL-IJCNLP5}, long-short-term-memory (RNN-LSTM) \citep{chen-EtAl:2015:EMNLP2} and convolutional neural network (CNN) \citep{Wang:2017} have been proposed. A transition score $[A]_{i, j}$ for jumping from $i$ to $j$ labels in successive characters is then introduced to handle the label-label transition and give a structured output. With the help of transition scores, a $|s| \times 4$ label score sequence can be decoded into a $|s| \times 1$ label sequence for inference, where $|s|$ is the number of the characters in a sentence. A post-processing module such as Conditional Random Fields (CRF) \citep{lafferty2001conditional} or maximum margin criterion \citep{taskar2004max} can be used to enforce structure consistency and provide an objective function.

	In word-based framework, \citet{zhang-zhang-fu:2016:P16-1} proposed a transition-based model which decodes a sentence from left-to-right incrementally.  \citet{cai-zhao:2016:P16-1} and \citet{cai-EtAl:2017:Short} proposed to score candidate segmented outputs directly. \citet{yang-zhang-dong:2017:Long} introduced partial-word into word-based models. For these word-based models, LSTMs \citep{hochreiter1997long} or their variants are used for feature extraction, maximum margin criterion is used for training and beam search is used for inference. 

	Despite of the great success these methods achieved, there are still problems in both character-based and word-based frameworks. The main problem of the character-based framework is the use of post-processing modules, e.g., CRF or maximum margin criterion. In computer vision literature, current state-of-the-art models tend to be in an end-to-end scheme and directly get output from the neural networks \citep{NIPS2015_5638, He_2017_ICCV}, which in comparison make these post-processing modules seem overdesigned.  Moreover, the use of \{B, I, E, S\} labels may produce redundant information. For example, ``B'' may be more similar to ``S'' or ``I'' than ``E''. These redundancies are not considered in all previous character-based models. An explanation we offer for the extensive using of these post-processing modules in the current state-of-the-art character-based models is that they are not good at capturing character combination features. Besides, the word-based models suffer from the problem of non-parallel, and they can only use the word segmentation information from the previous time steps.

	In this paper, we propose a concise and efficient approach that overcomes the problems of character-based and word-based models: To improve the feature combination, we introduce deep convolutional neural networks to extract features for segmentation. Moreover, we directly predict segmentation for each gap between two consecutive characters. Thus, we need not structure our scores and avoid inference decoding. Because technically speaking, our framework are segmenting based on the gaps, we refer to our approach as Gap-Based Convolutional Networks (Gap-Based ConvNets). Figure \ref{fig:framework} illustrates how our model works.

	We evaluate Gap-Based ConvNets on 5 different benchmark datasets, namely CTB6, PKU, MSR, AS and CityU. As a pure supervised model, Our approach outperforms the current state-of-the-art pure supervised results on all of these benchmarks by a large margin, while are also competitive with the best semi-supervised results. We hope that our simple framework will open up a new way and serve as a solid baseline for CWS research. We also hope that this paper can help ease future research in other sequence-labeling tasks. 
	
	The contributions of this paper could be summarized as follows.
\begin{itemize}
  \item We propose an end-to-end framework for CWS that directly classify the gaps between two consecutive characters and our results outperform the state-of-the-art character-based and word-based methods.
  \item Very deep neural networks are first introduced for CWS, in which we propose residual blocks and dense blocks to integrate multiple level character features. We also show that deeper neural networks can achieve better performance in CWS.
\end{itemize}

\section{Gap-Based Framework}
	Our gap-based framework is described  in detail in this section. 
	First of all, if we consider segmenting a given sentence $s$ (character sequence) into chunks (words), the most intuitive idea is to predict whether to segment for each gap between two consecutive characters. That is to say, $|s|-1$ predictions can determine the segmentation of the sentence $s$.

\subsection{Gap Feature Representation}
	
	We follow previous works and consider both uni-character embedding and bi-character embedding when transforming the one-hot sparse discrete sequence $s$ into real-valued representation.
	
	Two separated look-up tables are used to generate a 2-dimension representation $u$ of size $|s| \times e_u $ for uni-character embedding and a 2-dimension representation  $b$ of size $ ( |s| - 1 ) \times e_b$ for bi-character embedding, respectively, where $e_u$ is the dimension of uni-character embedding and $e_b$ is the dimension of bi-character embedding.
	
	The representation of a gap is then a concatenation of the uni-character embedding of its two consecutive characters and their bi-character embedding. Temporal convolutional layers (Conv) with kernel size 1 is applied to the concatenation, in order to integrate the uni-character and bi-character information. The output of the convolutional layer is regarded as the input to the feature extraction blocks.
	
	We follow \citet{gehring2017convolutional} and activate the convolutions with gated linear units (GLU) \citep{dauphin2016language}, which is a non-linearity operation that implement a gating mechanism over the convolution $Y = [A B]$:
	
	\begin{equation}
		v([A B]) = A \otimes \sigma(B)
	\end{equation}
	
Moreover, Batch normalization (BN) \citep{ioffe2015batch} follows each convolutional layer $A$ before it is scaled by the gate $\sigma(B)$. Biases are not used in convolutions.

	 In Figure \ref{fig:framework}, $u_i$ and $b_j$ denote the uni-character representation and the bi-character representation, respectively, and ``Concatenate'' denotes a concatenating operation. ``Conv-BN-GLU, k, d'' denotes a temporal convolutional operation of kernel size $k$ and kernel number $d$, which is then batch normalized and activated by GLU.
	 
\subsection{Feature Extraction}

Most of the previous applications of neural network to CWS use an architecture which is rather shallow (up to 5 layers). \citet{Wang:2017} proposed a 5-layer convolutional neural network. \citet{chen-EtAl:2015:EMNLP2} compared their LSTM models in different layers and found their 1-layer LSTM model works best. \citet{chen-EtAl:2017:Long2}, \citet{cai-EtAl:2017:Short}, \citet{yang-zhang-dong:2017:Long} and  \citet{zhou-EtAl:2017:EMNLP2017} also use 1-layer LSTM or bi-LSTM to extract features. These architectures are rather shallow in comparison to the deep convolutional networks which have pushed the state-of-the-art in computer vision. Besides, the use of $5$-character context window \citep{chen-EtAl:2017:Long2, zhou-EtAl:2017:EMNLP2017, yang-zhang-dong:2017:Long} in these models shows that their models are not good at capturing character combination features.

In this section, we propose deep feature extraction blocks, namely, residual blocks and dense blocks, to capture character combination features. To the best of our knowledge, this is the first time that very deep convolutional networks have been applied to sequence-labeling.

\paragraph{Residual Block}

\begin{figure}[t]
\includegraphics[width=7.7 cm]{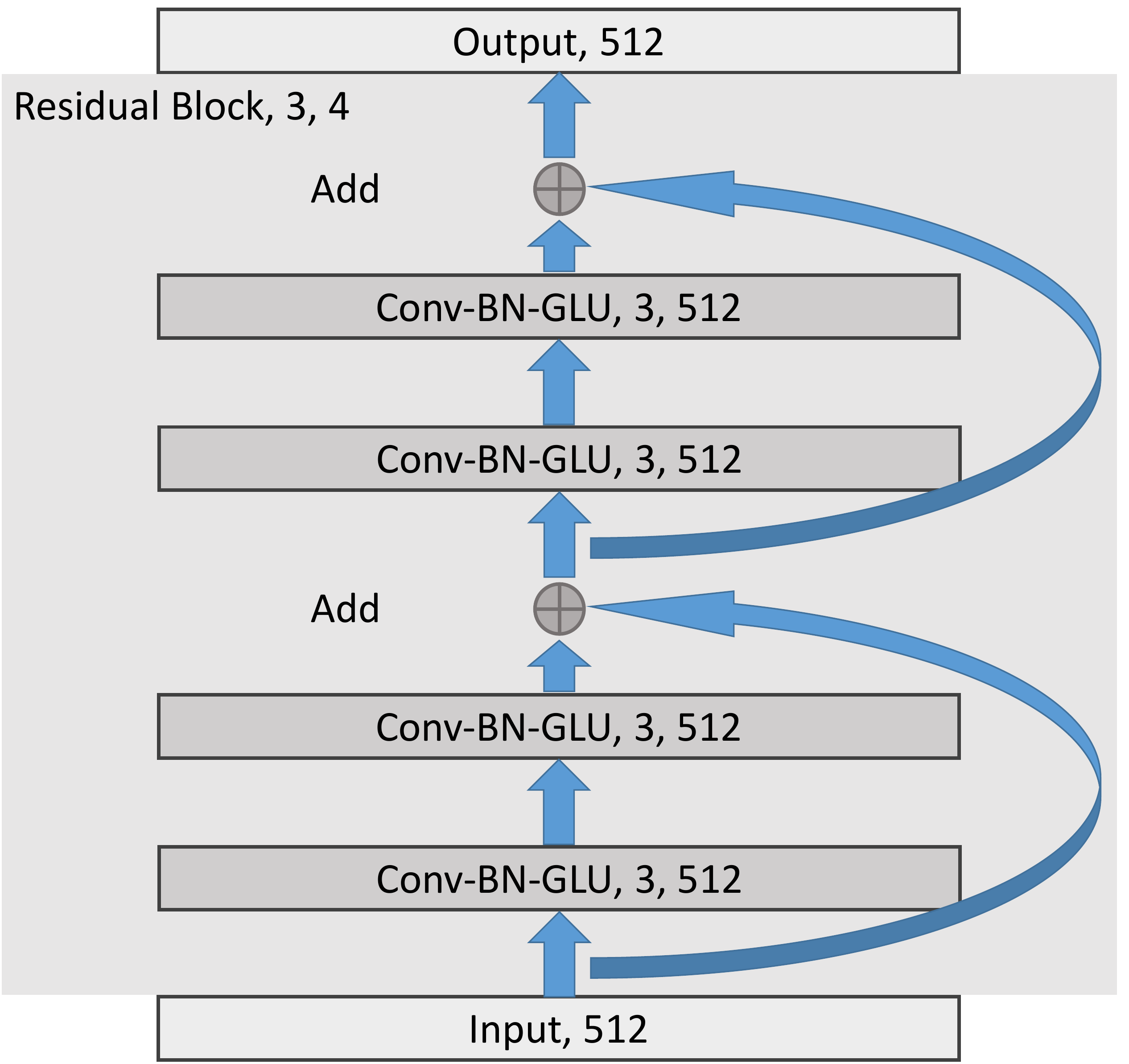}
 \begin{CJK}{UTF8}{gbsn}
\caption{A 4-layer residual block with convolutional kernel size 3, input depth and output depth 1024, denoted by ``Residual Block, 3, 4''.}
\label{fig:residualblock}
\end{CJK}
\end{figure}

	Traditional convolutional feed-forward networks connect the output of the  $\ell$-th layer as input to the $(\ell + 1)$-th layer. We can represent this scheme as a layer transition: $x_\ell = H_\ell(x_{\ell-1})$. Observing that deep feed-forward networks are hard to train, ResNets \citep{he2016deep} add a skip-connection that bypasses the non-linear transformations with an identity function:
	
	\begin{equation}
		x_\ell = H_\ell(x_{\ell-1}) + x_{\ell-1}
	\end{equation}
	
	Our $L$-layer wide residual block design follows \citet{zagoruyko2016wide}. We define $H_\ell(\cdot)$ as a composition of two gated linear unit (GLU), where each of them split the kernel-size-3 convolutions (Conv) into two parts, namely, $A$ and $B$, and control the convolutional layer $A$ with the gate $\sigma(B)$. Each $A$ is batch normalized (BN) before the gate.
	
	The Gap-Based ConvNets with a residual block is referred as Gap-Based ResNets in the rest of the paper. Figure \ref{fig:residualblock} illustrates the layout of an example residual block schematically. 
	
\paragraph{Dense Block}

\begin{figure}[t]
\includegraphics[width=7.7 cm]{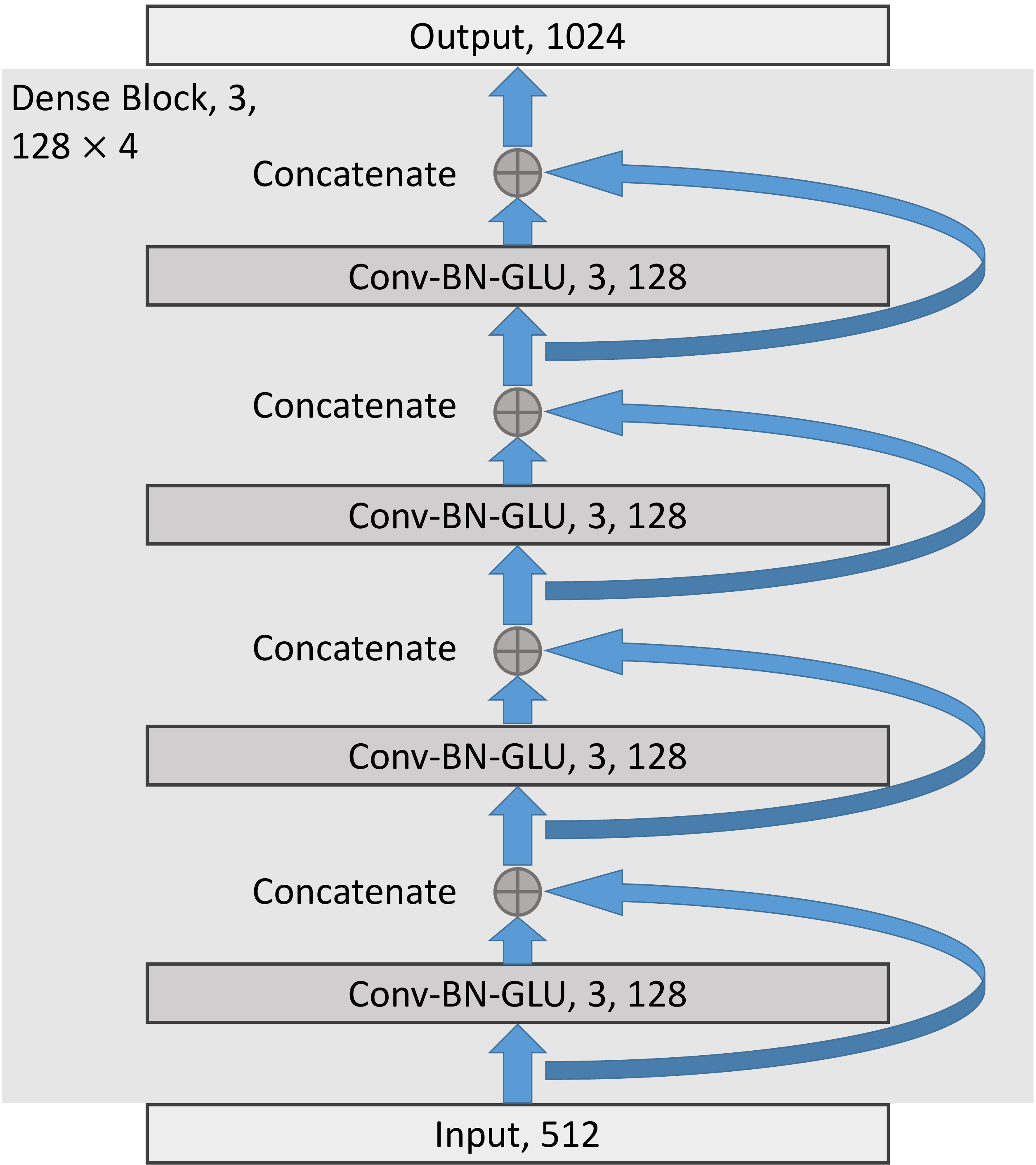}
 \begin{CJK}{UTF8}{gbsn}
\caption{A non-bottleneck 4-layer dense block with a growth rate of 128, convolutional kernel size 3, input depth 512 and output depth 512 + 128 $\times$ 4 = 1024, denoted by ``Dense Block, 3, 128 × 4''. Each layer takes the concatenation of all preceding feature-maps as input.}
\label{fig:denseblock}
\end{CJK}
\end{figure}
	
	The densely connected architecture \citep{Huang_2017_CVPR} is an extension of the residual architecture. Consider the gap representation $x_0$ are going to pass through the dense block and our dense block has $L$ layers. Then each layer implements a non-linear transformation $H_{\ell}(\cdot)$ to the concatenation of all preceding layers' feature maps:
	\begin{equation}
		x_\ell = H_\ell([x_0, x_1, \dots, x_{\ell-1}])
	\end{equation}
where $\ell$ indexes the layer and $[x_0, x_1, \dots, x_{\ell-1}]$ refers to the concatenation of the feature-maps produced in layers $0, \dots, \ell-1$.

	Similar to the residual blocks, we define $H_\ell(\cdot)$ as a gated linear unit (GLU) with a temporal convolutional layer (Conv) with kernel size 3 and Batch normalization (BN). 
	
	The number of the units in the temporal convolutional layer is referred as the growth rate $k$, as the concatenated layer depth grows by adding more layers. The Gap-Based ConvNets with such a dense block is referred as Gap-Based DenseNets.
	
	A temporal convolutional layer with kernel size 1 is introduced as the bottleneck layer before each convolutional layer with kernel size 3 to reduce the number of input feature maps. We set the number of kernel units in the bottleneck layers to $k$ as the same as the growth rate. Thus, our $H_\ell$ will be Conv1-BN-GLU-Conv3-BN-GLU.
	
	Figure \ref{fig:denseblock} illustrates the layout of an example non-bottleneck dense block schematically. It can be noticed that a dense block can explicitly concatenate multi-level character combination features.
	
\subsection{Segmentation Prediction}

	At the end of the deep feature extraction blocks, a temporal convolutional layer with kernel size 1 and unit number 2 is performed. Then, a softmax layer is attached to make the output a probability distribution. Finally we get a $(|s| -1) \times 2$ matrix for our prediction, and the values on the second dimension represent the prediction scores of ``segmentation'' and ``no segmentation'', respectively.
	
	We ensemble several separately trained models by taking the means of their prediction scores.

\section{Training}

	Because our model directly classifies the gaps into ``segmentation'' and ``no segmentation'', we can simply use the cross entropy (CE) as our loss function:
	\begin{equation}
		L = - \sum_{i=1}^{|s|-1}  \sum_{x \in \{0, 1 \}} p_t^{(i)} (x) \log p_m^{(i)} (x)
	\end{equation}
where $p_t^{(i)}$ is the true probability distribution for the $i$-th gap, which may be $(0, 1)$ for ``segmentation'' or $(1, 0)$ for ``no segmentation'', and $p_m$ is the probability distribution predicted by our model for the $i$-th gap, namely, the output of the softmax layer.

	Considering that there are plenty of annotation inconsistencies in the current datasets \citep{gong-EtAl:2017:EMNLP2017}, we use a label smoothing \citep{szegedy2016rethinking} with factor $\beta = 0.1$ to prevent overfit and boost the model robustness.

	We use Adam \citep{kingma2014adam} with a mini-batch size of $n$ to optimize model parameters, with an initial learning rate $\alpha_1 = 0.002$ in the first $8000$ steps and $\alpha_2 = 0.0002$ in the rest steps. Model parameters are initialized by normal distributions as \citet{glorot2010understanding} suggested. A dropout for the gap representation with dropout rate $p_1$ is used to reduce overfitting. We set pretrained bi-character embedding fixed and only fine-tune the pretrained uni-character embedding.
\section{Experiments}

\begin{table}[t]
\centering
 \begin{tabular}{|c c c c|} 
 \hline
 \textbf{Model Setting} & \textbf{P} & \textbf{R} & \textbf{F}\\
 \hline
 \multicolumn{4}{|c|}{Gap-Based ResNets }\\
 \hline
 $L$ = 4 & 96.1 & 95.9 & 96.0\\
 \hline
 $L$ = 12 & 96.2 & 96.1 & 96.1\\
 \hline
  \multicolumn{4}{|c|}{Gap-Based DenseNets}\\
 \hline
$L$ = \ \ 4, $k=128$ & 96.0 & 96.0 & 96.0\\
$L$ = 12, $k=128$ & \textbf{96.4} & 96.2 & \textbf{96.3}\\
$L$ = 12, $k=256$ & 96.3 & 96.2 & 96.2\\
$L$ = 28, $k=128$ & 96.2 & \textbf{96.4} & \textbf{96.3}\\
 \hline
 \end{tabular}
\caption{Results of model selection on CTB6.}
\label{tab:CTB6}
\end{table}

\subsection{Experimental Settings}

\paragraph{Data}
We use Chinese Treebank 6.0 (CTB6) (LDC2007T36) \citep{xue2005penn} as our main dataset. We follow the official document and split the dataset into training, development and test data. In order to verify the robustness of our model, we additionally evaluate our models on SIGHAN 2005 bakeoff \citep{emerson2005second} datasets, where we randomly split $10\%$ data from the training data as development data.

We replace all the punctuation with ``$<$PUNC$>$'', English characters with ``$<$ENG$>$'' and Arabic numbers with ``$<$NUM$>$'' for all text. We also add ``$<$/s$>$'' symbol to the beginning and the end of a sentence.

We apply word2vec \citep{mikolov2013distributed} on Chinese Gigaword corpus (LDC2011T13) to get pretrained embedding of uni-characters and bi-characters. We choose $50$ for both uni-character embedding size $e_u$ and bi-character embedding size $e_b$. Notice that as we do not need word embedding, we do not have to automatically segment the corpus by other segmentors. The use of word embedding in the gap-based framework is left for future works.

\paragraph{Evaluation} 
The standard word precision, recall and F1 measure \citep{emerson2005second} are used to evaluate segmentation performances. The prediction scores of 4 separately trained models with same settings are ensembled for the evaluation of different model architectures and hyper-parameters.

\paragraph{Fine-tune} 
We fine-tune the hyper-parameters on the development data. We almost keep all the hyper-parameters to be the same when we evaluate our models on different datasets, except that the batch size is set to 256 for AS dataset while to 64 for other datasets, and  dropout rate $p_1$ is set to 0.3 for CTB6 and PKU datasets while to 0.2 for other datasets.

\subsection{Model Analysis}
We perform development experiments on CTB6 dataset to verify the usefulness of various configurations and different loss objectives, respectively.

\subsubsection{Model Selection}

We evaluate our Gap-Based Convolutional Networks with residual blocks or dense blocks, with different number of layers $L$. The main results on CTB6 are shown in Table \ref{tab:CTB6}, where we mark our best results in \textbf{boldface}. The dimension of the first convolutional layer (gap representation) are set to 512.

We find that our $12$-layer and $28$-layer Gap-Based DenseNets both achieve the best performance, with growth rate 128. Therefore, to get both the performance and the speed, we only use a $12$-layer Gap-Based DenseNet with growth rate 128 in the following experiments.

\subsubsection{Gap Representation}

\begin{table}[t]
\centering
 \begin{tabular}{|c c c c|} 
 \hline
 \textbf{Context} & \textbf{P} & \textbf{R} & \textbf{F}\\
 \hline
 both & \textbf{96.4} & \textbf{96.2} & \textbf{96.3}\\
 \hline
 bi-character & 94.9 & 94.6 & 94.8\\
 \hline
 uni-character & 95.8 & 95.8 & 95.8\\
 \hline
 oracle-combined & \textbf{97.4} & \textbf{97.3} & \textbf{97.4}\\
 \hline
 \end{tabular}
\caption{Influence of different gap representation.}
\label{tab:context}
\end{table}

We compare different gap representations. The results are shown in Table \ref{tab:context}, where ``both'' represents the original model, ``uni-character' ' and ``bi-character' ' represent the gap representation only with uni-character embedding and bi-character embedding, respectively. And ``combined'' represents the combined results of a pure uni-character model and a pure bi-character model.

As can be seen from the table, by removing uni-character and bi-character embedding, the
F-score decreases to $94.8$ and $95.8$, respectively. We can find that the uni-character embedding are more robust and useful than bi-character embedding, which is an opposite conclusion to \citet{yang-zhang-dong:2017:Long}. We believe this is due to the sparsity of the bi-character embedding.

\begin{table}[t]
\centering
\small
 \begin{CJK}{UTF8}{gbsn}
 \begin{tabular}{|c c c|}
 \hline
 \textbf{Context} &  \textbf{Sample} &  \textbf{Correct}\\
 \hline
 uni-character & \tabincell{c}{他 才 又 有 机会 站到\\ 火车 修复 的 第一 线} & $\surd$\\
 \hline
 bi-character & \tabincell{c}{他 才 又 有 机会 站 到\\ 火车修 复 的 第一 线}& $\times$\\
 \hline
 uni-character & \tabincell{c}{美 不 胜收 的 阿里山} & $\times$\\
 \hline
 bi-character & \tabincell{c}{美不胜收 的 阿里山}& $\surd$\\
 \hline
 \end{tabular}
  \end{CJK}
\caption{
 \begin{CJK}{UTF8}{gbsn} Example segmentation results of different character contexts. The correct segmentation should be ``他(he) 才(just) 又(again) 有(have) 机会(chance) 站到(stand) 火车(train) 修复(repair) 的('s) 第一(first) 线(frontier)'' and ``美不胜收(beautiful) 的(of) 阿里山(Ali Mountain)''.
 \end{CJK}
}
\label{tab:context_sample}
\end{table}

In addition, we conduct a simple experiment on the combination of pure uni-character and pure bi-character models.

As the pure uni-character models and the pure bi-character models have different representations for the gaps, their predictions for the segmentation are also independent. They produce different distribution of segmentation errors, which provide the opportunity for them to learn from each other, as shown in Table \ref{tab:context_sample}.

Therefore, we provide an oracle-combined model to combine the results from these two models. For the combined results, we accept the segmentation results that both pure uni-character model and pure bi-character model accept and use what we call ``oracle'' to determine whose results to accept when in divergence of views. The ``oracle'' represents that we can always make the right choices. The combined results are also listed in Table \ref{tab:context}. 

We find that while performs better than pure uni-character and pure bi-character models, our oracle-combined results also outperform the original model by a large margin, which suggests that uni-character information and bi-character information are quite complementary, and maybe the combination of uni-character information and bi-character information before the feature extraction blocks makes the original model not able to sufficiently exploit the feature combination ability of deep neural networks.

However, ``oracle'' means that we need to know the answer beforehand, which is impossible in practice. Therefore, a practical way to combine the results of pure uni-character and pure bi-character models is left for future investigation.

\subsubsection{Comparison with Character-Based Framework}

Both our gap-based model and character-based models regard CWS task as a sequence-labeling task. However, a big difference between them is that while character-based models have to give 4 scores for each character and use a post-processing module, the gap-based models only need to give a binary classification for each gap. 

In order to see the efficiency of the gap-based schemes, we compare the character-based scheme and the gap-based scheme in the same neural network architecture, namely, Gap-Based DenseNets. To combine the character-based scheme with the original architecture, we revise the architecture by
\begin{itemize}
  \item We represent the character information by the character's uni-character embedding and the bi-character embedding of its consecutive characters.
  \item We use a convolutional layer and softmax layer with $4$ units instead of $2$, which represent the scores for \{B, I, E, S\} in the character-based sequence-labeling scheme. 
\end{itemize}

\begin{table}[t]
\centering
 \begin{tabular}{|c c c c|} 
 \hline
 \textbf{Scheme} & \textbf{P} & \textbf{R} & \textbf{F}\\
 \hline
 gap-based & \textbf{96.4} & \textbf{96.2} & \textbf{96.3}\\
 \hline
 character-based CRF & 96.1 & 96.3 & 96.2\\
 \hline
 character-based greedy & 96.1 & 96.1 & 96.1\\
 \hline
 \end{tabular}
\caption{Influence of different CWS frameworks.}
\label{tab:scheme}
\end{table}

\begin{table}[t]
\centering
\small
 \begin{CJK}{UTF8}{gbsn}
 \begin{tabular}{|c c c|}
 \hline
 \textbf{Framework} &  \textbf{Sample} &  \textbf{Correct}\\
 \hline
 gap-based & \tabincell{c}{中国 证券 市场 目前\\ 仍 处 于 初创 阶段} & $\surd$\\
 \hline
 character-based & \tabincell{c}{中国 证券 市场 目前\\  仍 处于 初 创 阶段}& $\times$\\
 \hline
 \end{tabular}
  \end{CJK}
\caption{
 \begin{CJK}{UTF8}{gbsn}
An example segmentation result in different frameworks. The correct segmentation should be ``中国(Chinese) 证券(stock) 市场(market) 目前(currently) 仍(still) 处(in) 于(the) 初创(start) 阶段(period)''.
 \end{CJK}}

\label{tab:framework_sample}
\end{table}

\begin{table*}[t]
\centering
 \begin{tabular}{|c|c|c|c|c|c|c|} 
 \hline
\textbf{F1 score} & \textbf{CTB6} & \textbf{PKU} & \textbf{MSR} & \textbf{AS} &\textbf{CityU}\\
 \hline
 our proposed & \textbf{96.3} & \textbf{96.0} & \textbf{97.9} & \textbf{96.1} & \textbf{96.9}\\
 \hline
 \hline
 \citet{cai-EtAl:2017:Short} &- & 95.8 & 97.1 & 95.6 & 95.3\\
 \hline
 \citet{zhou-EtAl:2017:EMNLP2017} baseline &94.9& 95.0 & 97.2 & - & -\\
 \hline
 \citet{Wang:2017} W2VBE-CONV &-& 95.9 & 97.5 & - & -\\
 \hline
 \citet{cai-zhao:2016:P16-1} & - & 95.5 & 96.5 & - & -\\
 \hline
 \citet{chen-EtAl:2015:ACL-IJCNLP5} GRNN* & - & 94.5 & 95.4 & - & -\\
 \hline
 \citet{chen-EtAl:2015:EMNLP2} LSTM* & - & 94.8 & 95.6 & - & -\\
 \hline
 \citet{wang-voigt-manning:2014:P14-2} dual &-& 95.3 & 97.4 & 95.4 & 94.7\\
 \hline
 \citet{sun:2010:POSTERS} & - & 95.2 & 96.9 & 95.2 & 95.6\\
 \hline
 \citet{zhang-clark:2007:ACLMain} &-& 94.5 & 97.2 & 94.6 & 95.1\\
 \hline
 \end{tabular}
\caption{Main results on CTB6 and SIGHAN 2005 bakeoff datasets with other best pure supervised results. Results with * are from the implementation of \citet{cai-zhao:2016:P16-1}, which do not use an external dictionary of Chinese lexicons\citep{chen-EtAl:2015:ACL-IJCNLP5, chen-EtAl:2015:EMNLP2}.}
\label{tab:results}
\end{table*}

\begin{table*}[t]
\centering
 \begin{tabular}{|c|c|c|c|c|c|c|} 
 \hline
\textbf{F1 score} & \textbf{CTB6} & \textbf{PKU} & \textbf{MSR} & \textbf{AS} &\textbf{CityU}\\
 \hline
 our proposed & \textbf{96.3} & 96.0 & 97.9 & \textbf{96.1} & \textbf{96.9}\\
 \hline
 \hline
 \citet{zhou-EtAl:2017:EMNLP2017} WCC embeddings\dag & 96.2 & 96.0 & 97.8 & - & -\\
 \hline
 \citet{yang-zhang-dong:2017:Long} multi-pretrain* & 96.2 & 96.3 & 97.5 & 95.7 & 96.9\\
 \hline
 \citet{Wang:2017} WE-CONV* & - & \textbf{96.5} & \textbf{98.0} & - & -\\
 \hline
 \citet{zhang-zhang-fu:2016:P16-1} neural* &95.0 & 95.1 & 97.0 & - & -\\
 \hline
 \citet{zhang-zhang-fu:2016:P16-1} hybrid* & 96.0 & 95.7 & 97.7 & - & -\\
 \hline
 \end{tabular}
\caption{Main results on CTB6 and SIGHAN 2005 bakeoff datasets with other best semi-supervised results. The results marked with * and \dag use auto-segmented data to get pretrained word embeddings and character embeddings, respectively.}
\label{tab:semi-results}
\end{table*}

To fully investigate the performance of character-based scheme, following \citep{zhou-EtAl:2017:EMNLP2017}, we train the network in two approaches. The first approach use a transition matrix to model the tag dependency and CRF for structured inference, and the second use a greedy loss that directly classifies the characters into \{B, I, E, S\}. The results are shown in Table \ref{tab:scheme}, where ``character-based CRF'' represents the first approach and ``character-based greedy'' represents the second.

We can see that our gap-based scheme have competitive results to the character-based scheme, while we do not have a post-processing module. Moreover, our simple classification scheme makes us more convenient to use a great deal of well-developed tricks in the large supervised learning literature, e.g., label smoothing and confidence penalty.

Table \ref{tab:framework_sample} gives an example segmentation result that our gap-based framework gives the correct segmentation, while the answer from the character-based framework is wrong. It shows that our gap-based segmentation framework is able to fix some mistakes that character-based framework will make.

\subsection{Final Results}

In addition to CTB6 dataset, which has been the most commonly adopted by recent segmentation research, we additionally evaluate our models on the SIGHAN 2005 bakeoff datasets, to examine cross domain robustness. Among these datasets, PKU and MSR datasets are in simplified Chinese, while AS and CityU datasets are in traditional Chinese and we have to map them into simplified Chinese before segmentation.

Our final results are shown in Table \ref{tab:results} and Table \ref{tab:semi-results}, which list the results of several current state-of-the-art methods. As can be seen from Table \ref{tab:results}, our proposed method gives the best pure supervised performance among both statistical and neural segments in all datasets by a large margin.

Moreover, our method are also competitive to the best semi-supervised methods, including those using rich pretraining information like mutual information\citep{sun-xu:2011:EMNLP}, punctuation\citep{sun-xu:2011:EMNLP, yang-zhang-dong:2017:Long}, automatically segmented text\citep{zhou-EtAl:2017:EMNLP2017, yang-zhang-dong:2017:Long}, POS data\citep{sun-xu:2011:EMNLP, yang-zhang-dong:2017:Long} or word-context embedding\citep{zhou-EtAl:2017:EMNLP2017}. As can be seen in Table \ref{tab:semi-results}, we outperform the best semi-supervised models on CTB6, AS and CityU datasets.

In summary, while competitive to the best semi-supervised methods, our gap-based approach gives the best pure supervised performance on all corpora. To our knowledge, we are the first to report state-of-the-art results on both CTB6 and all SIGHAN 2005 bakeoff benchmarks. It verifies that while simple, the gap-based framework is very efficient for CWS.

\section{Related Work}

\citet{O03-4002} was the first to propose to regard CWS task as character-tagging, using a maximum entropy model to give each character a label from  \{B, I, E, S\}. \citet{peng2004chinese} followed this character-tagging scheme and proposed a conditional random field (CRF) to further improve the performance. Since then, this sequence-labeling CRF scheme was followed by most subsequent approaches in the literature.

\citet{zheng2013deep} was the first to propose a neural network model and introduced character embedding for CWS, using a character window of size 5. This kind of 5-character window design still appears in recent state-of-the-art models. \citet{pei-ge-chang:2014:P14-1} used a tensor neural network to further exploit the character combination feature and introduced bi-character embedding into neural CWS models. \citet{chen-EtAl:2015:ACL-IJCNLP5} proposed a Gated Recursive Neural Network to combine character features. \citet{chen-EtAl:2015:EMNLP2} proposed a LSTM model in order to get rid of this 5-character window design. \citet{xu-sun:2016:P16-2} proposed a Dependency-based Gated Recursive Neural Network to efficiently combine local and global features. \citet{Wang:2017} proposed a convolutional neural network to extract features. These character-based models exploited the recent springing up neural network architectures.

\citet{a-hybrid-markovsemi-markov-conditional-random-field-for-sequence-segmentation} introduced the log conditional odds that a given token sequence constitutes a chunk according to a generative model. \citet{zhang-clark:2007:ACLMain} was the first to proposed a word-based approach to CWS, which provides a direct solution to the problem. \citet{zhang2011syntactic} proposed a beam-search model. \citet{zhang-zhang-fu:2016:P16-1} proposed a neural transition-based model with beam search that explicitly produce chunks in order. \citet{cai-zhao:2016:P16-1} and \citet{cai-EtAl:2017:Short} proposed a model that score the candidate segmented outputs directly. These word-based models can fully exploit the word features such as word embedding.

Observing the similarity of sequence segmentation in natural language processing (NLP) literature and and semantic segmentation in computer vision literature, we may find that ``simplest things are the best'' Occam's razor is not applied to NLP sequence segmentation tasks. \citet{zhang-zhang-fu:2016:P16-1} proposed to use Fully Convolutional DenseNets to do semantic segmentation by directly do  classification on each pixel and beat other complex framework such as fully connected CRFs \citep{krahenbuhl2011efficient}. This paper provides us the idea to directly classify the gaps.

The model architecture of \citet{Wang:2017} is similar to ours, as both of us use convolutional neural networks to extract character combination features. However, their models are rather shallow (up to 5 layers) and only use feed forward connections, while we introduce deep feature extraction blocks that contain residual connections or dense connections, inspired by \citet{he2016deep} and \citet{Huang_2017_CVPR}. Moreover, their models are still in character-based framework, while we show that our gap-based framework can further exploit the representation power of deep neural networks.

Our approach remains much potential that can be further investigated and improved in the future. For example, our models may furthermore benefit from recently popular semi-supervised learning methods, such as word-context character embedding\citep{zhou-EtAl:2017:EMNLP2017}, rich pretraining\citep{yang-zhang-dong:2017:Long} and pretrained word embedding \citep{Wang:2017}. They can all get more information from auto-segmented text. The use of LSTM-RNN and its variants in the gap-based framework is also interesting to be investigated.

\section{Conclusion}
In this paper, we propose a novel gap-based framework for Chinese word segmentation that directly predict whether to segment for each gap between two consecutive characters. Moreover, we introduce very deep convolutional networks (residual blocks and dense blocks) for feature extraction.

Experiments show that our proposed Gap-Based ConvNets are effective to solve Chinese word segmentation task. We outperform the previous best character-based and word-based methods by a large margin. To our knowledge, we are the first to report state-of-the-art results on both CTB6 and all SIGHAN 2005 bakeoff benchmarks.

\bibliography{CWS}
\bibliographystyle{acl_natbib}

\end{document}